\documentclass[letterpaper, 10pt, journal, twoside, final]{IEEEtran}

\usepackage[utf8]{inputenc}
\usepackage[T1]{fontenc}
\usepackage{acro}
\usepackage{algorithmicx, algpseudocode}
\usepackage{amsfonts}
\usepackage{amsmath}
\usepackage{amssymb}
\usepackage{cite}
\usepackage[keeplastbox]{flushend}
\usepackage{pgfplots}
\usepackage[separate-uncertainty=true]{siunitx}
\usepackage[caption=false,font=footnotesize]{subfig}  
\usepackage{tikz}
\usepackage{url}

\DeclareAcronym{BCE}{short=BCE, long=binary crossentropy}
\DeclareAcronym{DoF}{short=DoF, long=degrees of freedom}
\DeclareAcronym{FCNN}{short=FCNN, long=fully-convolutional neural network, short-plural-form=FCNNs}
\DeclareAcronym{MDP}{short=MDP, long=Markov decision process}
\DeclareAcronym{NCE}{short=NCE, long=noise contrastive estimation}
\DeclareAcronym{NN}{short=NN, long=neural network}
\DeclareAcronym{RL}{short=RL, long=reinforcement learning}
\DeclareAcronym{TCP}{short=TCP, long=tool center point}

\DeclareMathOperator*{\argmax}{arg\,max}

\algnewcommand\algorithmicdata{\textbf{Data:}}
\algnewcommand\Data{\item[\algorithmicdata]}
\algnewcommand\algorithmicresult{\textbf{Result:}}
\algnewcommand\Result{\item[\algorithmicresult]}

\pgfplotsset{compat=1.14}
\usetikzlibrary{backgrounds, calc}
\usepgfplotslibrary{groupplots}
\newcommand*\circled[1]{\tikz[baseline=(char.base)]{\node[shape=circle, draw, inner sep=2pt, minimum size=14pt] (char) {#1};}}

\begin{document}

\title{Self-supervised Learning for Precise\\ Pick-and-place without Object Model}

\author{Lars Berscheid, Pascal Meißner, and Torsten Kröger%
    \thanks{Manuscript received: February, 23, 2020; Revised May, 13, 2020; Accepted June, 5, 2020.}
    \thanks{This paper was recommended for publication by Editor  Hong Liu upon evaluation of the Associate Editor and Reviewers' comments.}
    \thanks{The  authors  are  with  the Intelligent Process Automation and Robotics Lab (IPR), Karlsruhe Institute of Technology (KIT), Karlsruhe, Germany {\tt\footnotesize \{lars.berscheid, pascal.meissner, torsten\}@kit.edu}}%
    \thanks{Digital Object Identifier (DOI): see top of this page.}
}

\markboth{IEEE Robotics and Automation Letters. Preprint Version. Accepted June, 2020}
{Berscheid \MakeLowercase{\textit{et al.}}: Self-supervised Learning for Pick-and-place}

\maketitle

\begin{abstract}
Flexible pick-and-place is a fundamental yet challenging task within robotics, in particular due to the need of an object model for a simple target pose definition. In this work, the robot instead learns to pick-and-place objects using planar manipulation according to a single, demonstrated goal state. Our primary contribution lies within combining robot learning of primitives, commonly estimated by fully-convolutional neural networks, with one-shot imitation learning. Therefore, we define the place reward as a contrastive loss between real-world measurements and a task-specific noise distribution. Furthermore, we design our system to learn in a self-supervised manner, enabling real-world experiments with up to \num{25000} pick-and-place actions. Then, our robot is able to place trained objects with an average placement error of \SI{2.7 \pm 0.2}{mm} and \SI{2.6 \pm 0.8}{^\circ}. As our approach does not require an object model, the robot is able to generalize to unknown objects while keeping a precision of \SI{5.9 \pm 1.1}{mm} and \SI{4.1 \pm 1.2}{^\circ}. We further show a range of emerging behaviors: The robot naturally learns to select the correct object in the presence of multiple object types, precisely inserts objects within a peg game, picks screws out of dense clutter, and infers multiple pick-and-place actions from a single goal state.
\end{abstract}

\section{Introduction}

\IEEEPARstart{T}{he} task of grasping and placing an object with desired accuracy is essential for robotic object handling in general, and even more so for today's industrial and logistics automation \cite{siciliano_springer_2016}. The ideal solution to this so-called \textit{pick-and-place} task needs to fulfill a wide range of requirements: First, 
for greatest flexibility the approach should even work for unknown objects without model. Second, it needs to ensure high reliability for picking objects out of dense clutter or an obstacle-rich environment like a randomly filled bin. Third and important for real-world applications, the computation needs to be as fast as possible, all while keeping the desired placing precision.

Robot learning has shown significant progress in recent years, enabling skills like grasping of unknown objects or pre-grasping manipulation. For many real-world use cases, these approaches have improved the flexibility and reliability of grasping significantly. However, transferring these approaches from grasping alone to pick-and-place yields two essential challenges:
First, grasping and placing are interdependent; influencing and limiting each other in cluttered scenarios or for high-accuracy requirements.
Second, how does the robot know where to place a never-seen object? We will address this question by using a single demonstration as a goal state, leading to an approach known as \textit{one-shot imitation learning}.

In this work, we emphasize the precision of pick-and-place and define our desired accuracy in the order of millimeters. In contrast, if the object sizes exceed the desired precision of the task, we found the term \textit{pick-and-place} to be used as well \cite{zeng2018robotic, finn2017one}. Then, the interdependence between grasping and placing can be neglected, simplifying the problem tremendously. Moreover, we restrict ourselves to the following constraints: First, we limit the robot to planar manipulation. Second, we separate the grasping and placing scene spatially, allowing for more industrial use-cases. Third, we consider multiple object types in the grasping scene. The robot then needs to select and grasp the demonstrated object. 

\begin{figure}[t]
	\center
\begin{tikzpicture}
    \node[anchor=south west, inner sep=0] (overall) at (0,0) {\includegraphics[trim=20 50 20 60, clip, width=0.78\linewidth]{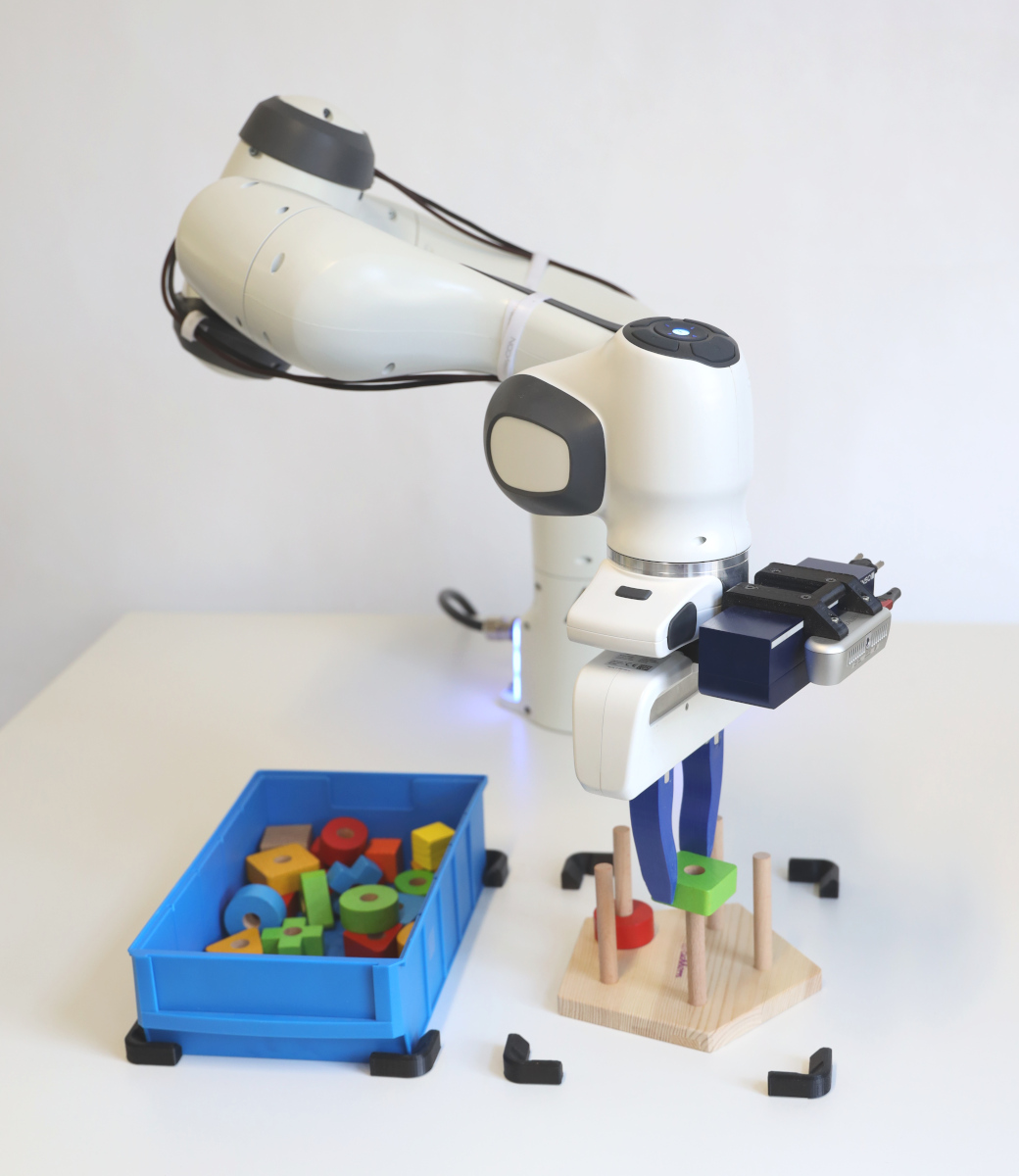}};
	\begin{scope}[x={(overall.south east)}, y={(overall.north west)}]
		\node[] at (0.88, 0.82) {\circled{1}};
		\draw [->] (0.855, 0.80) to (0.68, 0.63);
		
		\node[] at (0.88, 0.67) {\circled{2}};
		\draw [->] (0.867, 0.643) to (0.78, 0.45);
		
		\node[] at (0.11, 0.55) {\circled{3}};
		\draw [->] (0.141, 0.535) to (0.67, 0.36);
		
		\node[] at (0.11, 0.4) {\circled{4}};
		\draw [->] (0.135, 0.383) to (0.32, 0.25);
	\end{scope}
\end{tikzpicture}

	\caption{Our robot is able to grasp objects out of clutter and place them with high precision, given a (demonstrated) goal state. The system has learned in a self-supervised manner with minimal human interaction in the real-world and does not depend on an object model, enabling the robot to generalize to unknown objects. We use a robot arm (1), depth cameras (2), planar manipulation with a two-finger gripper (3) and bins with various objects (4).}
	\label{fig:overall-system}
\end{figure}
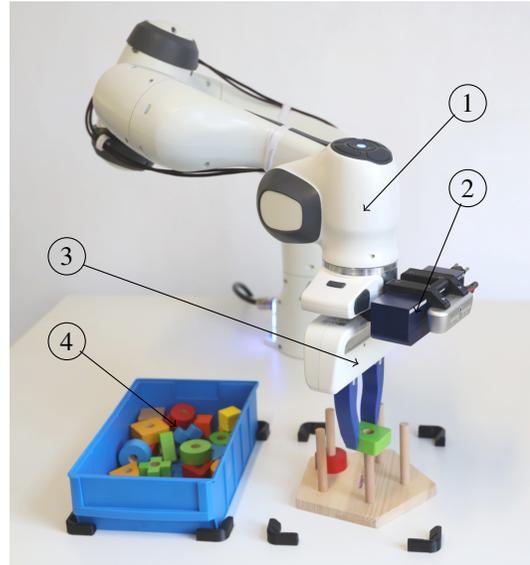

We see our contributions as follows: First, we extend our approach of learning using manipulation primitives and \acp{FCNN} for reward estimation to pick-and-place actions. Second and most importantly, we combine this approach with one-shot imitation learning for defining the goal pose flexibly. Third, we design a self-supervised learning strategy without human interaction, enabling the real-world training to scale to over \num{120} robot hours. Finally, we evaluate our system for pick-and-place precision and selection accuracy quantitatively, and demonstrate its limits in a range of tasks qualitatively. This includes placing screws out of dense clutter in an industrial use case, as well as inferring multiple actions from a single demonstration.

\section{Related Work}

Object handling, and in particular grasping as a first interaction for further manipulation, are well-researched areas within robotics. Regarding grasping, Bohg et al.~\cite{bohg_data-driven_2014} differentiate between analytical and data-driven approaches. Historically, grasps were synthesized commonly based on analytical constructions of force-closure \cite{ferrari_planning_1992}. In comparison, data-driven approaches sample grasps using object recognition, pose estimation or specific feature extraction \cite{bohg_data-driven_2014, miller_graspit!_2004}. For known objects, pick-and-place reduces to estimating the object's pose, a grasp point, \textit{and} the following pose displacement during the grasping process. In the following however, we will focus on the case of manipulation without object model.

In recent years, manipulation has seen great progress as a key area in robot learning. From our point of view, two fundamental approaches have emerged: First, an \textit{end-to-end} approach using a step-wise, velocity-like control of the end effector \cite{duan2017one, kalashnikov_2018_qt, quillen_deep_2018}. Second, with the usage of predefined \textit{manipulation primitives} a controller needs to decide where to apply which primitive. This is often combined with a \ac{FCNN}, estimating the reward for a discrete set of poses in the image space. Grasping can then be learned in simulation like Dex-Net \cite{mahler_dex-net_2017}, with real-world interaction \cite{berscheid_improving_2019}, or including pre-grasping manipulation for dense clutter \cite{zeng_learning_2018, berscheid_shifting_2019}. 

\textit{Task-based grasping} investigates the interdependence between grasps and the subsequent task \cite{li2007data, bohg2012task, pardi2018choosing}. Recently, the TossingBot~\cite{zeng2019tossingbot} has learned to pick and throw objects into target bins, extending prior work of primitive-based grasping. However, future toss actions do not influence prior grasps.

Regarding pick-and-place itself, Zeng et al.~\cite{zeng2018robotic} used classical image matching to pick-and-place in the broader sense of \textit{semantic grasping}, without further requirements for a precise place pose. While the robot was able to grasp objects out of clutter reliably, grasping was trained in a human-annotated manner.
Gualtieri et al.~\cite{gualtieri2018pick} learned pick-and-place of 6-\ac{DoF} in simulation and were able to transfer the results to real-world cluttered scenes. However, the robot is limited in generalizing to unknown object classes and novel place poses. Besides grasping, some work has focused on individual subproblems of pick-and-place. Jiang et al.~\cite{jiang2012learning} have focused on finding a stable place pose given a grasping configuration. Zhao et al.~\cite{zhao2019towards} calculated object displacements during the grasping action without an object model.

Pick-and-place is a popular task within imitation and \ac{RL}. Finn et al.~\cite{finn2017one} used meta-learning on multiple pick-and-place tasks to achieve one-shot imitation learning for novel objects and scenarios. Their accuracy suffices for placing objects within a larger box or bowl, however their grasping approach seems restricted to non-cluttered scenarios.
Duan et al.~\cite{duan2017one} used one-shot imitation learning for the task of block stacking, including training in simulation, virtual reality, and a final sim-to-real transfer.
Singh et al.~\cite{singh2019end-to-end} has learned robot manipulation in the context of Inverse \ac{RL}. While their bookshelf scenario is quite simplified in the context of pick-and-place, their method allows for an easy and flexible definition of a goal state without explicit reward definition.

\section{Learning for Pick-and-place}

We introduce our system using the notation of \ac{RL}, however limited to a single action step. Then, the underlying \ac{MDP} is defined by its tuple $(\mathcal{S}, \mathcal{A}, r)$ with the state space $\mathcal{S}$, the action space $\mathcal{A}$, and the reward function $r$. Furthermore, we will specify the action space (\ref{subsec:manipulation-primitives}), the state space observations (\ref{subsec:state-space}), and the learned reward function (\ref{subsec:reward-function}) in detail. The solution to the \ac{MDP} is a policy $\pi: \mathcal{S} \rightarrow \mathcal{A}$ mapping the current state $s_t \in \mathcal{S}$ to an action $a_t \in \mathcal{A}$.

\subsection{Manipulation Primitives}
\label{subsec:manipulation-primitives}

A pick-and-place action $a \in \mathcal{A}_g \times \mathcal{A}_p$ is a \textit{grasp} $a_g \in \mathcal{A}_g$ following a \textit{place} action $a_p \in \mathcal{A}_p$. We limit both action types to a set of manipulation primitives, given by predefined motions at a specified pose. Due to planar manipulation, each action space $\mathcal{A}_g$, $\mathcal{A}_p$ is given by $\text{SE}(2) \times \mathbb{R} \times \mathbb{N}$ with actions parametrized by a tuple $(x, y, \theta, z, i)$ with the planar translation $(x, y)$ parallel to the table surface, the 2D rotation $\theta$ around the $z$-axis and the index of the manipulation primitive $i$. The height $z$ is calculated directly from the depth image. In general, the controller needs to decide where to apply which manipulation primitive $i$. 

\begin{figure}[t]
	\centering
	\vspace{0.8mm}
\begin{tikzpicture}[
    image/.style={anchor=north west, inner sep=0, node distance=62},
    labels/.style={node distance=43, text width=52, align=center, font=\small\linespread{0.8}\selectfont}
]
    \node[image] (1-grasp-v) at (0, 0) {\includegraphics[width=0.23\linewidth]{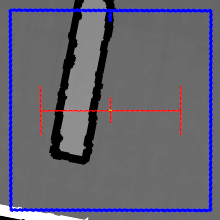}};
    
    \node[image, right of=1-grasp-v] (1-place-v) {\includegraphics[width=0.23\linewidth]{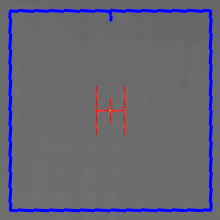}};
    
    \node[image, right of=1-place-v] (1-place-goal) {\includegraphics[width=0.23\linewidth]{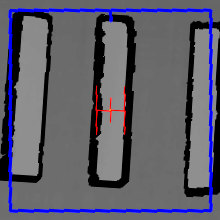}};
    
    \node[image, right of=1-place-goal] (1-place-after) {\includegraphics[width=0.23\linewidth]{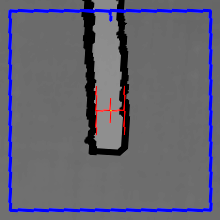}};
    
    \node[image, below of=1-grasp-v, node distance=65] (2-grasp-v) {\includegraphics[width=0.23\linewidth]{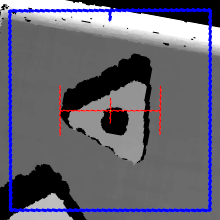}};
    
    \node[image, right of=2-grasp-v] (2-place-v) {\includegraphics[width=0.23\linewidth]{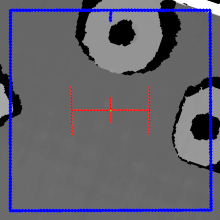}};
    
    \node[image, right of=2-place-v] (2-place-goal) {\includegraphics[width=0.23\linewidth]{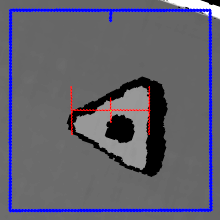}};
    
    \node[image, right of=2-place-goal] (2-place-after) {\includegraphics[width=0.23\linewidth]{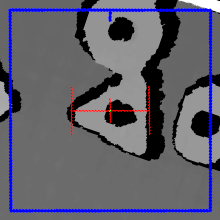}};
    
    \node[image, below of=2-grasp-v, node distance=65] (3-grasp-v) {\includegraphics[width=0.23\linewidth]{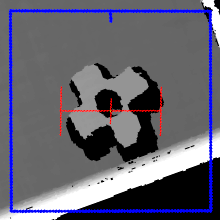}};
    
    \node[image, right of=3-grasp-v] (3-place-v) {\includegraphics[width=0.23\linewidth]{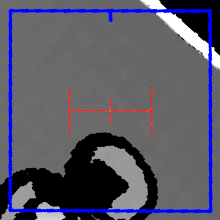}};
    
    \node[image, right of=3-place-v] (3-place-goal) {\includegraphics[width=0.23\linewidth]{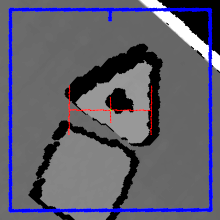}};
    
    \node[image, right of=3-place-goal] (3-place-after) {\includegraphics[width=0.23\linewidth]{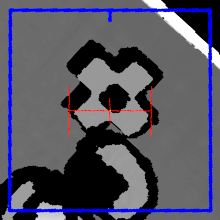}};

    \draw[] ($(1-grasp-v.north west) + (-0.05, 0.86)$) -- ($(1-place-after.north east) + (0.05, 0.86)$);
    \draw[] ($(1-grasp-v.north west) + (-0.05, 0.13)$) -- ($(1-place-after.north east) + (0.05, 0.13)$);
    \draw[] ($(2-grasp-v.north west) + (-0.05, 0.13)$) -- ($(2-place-after.north east) + (0.05, 0.13)$);
    \draw[] ($(3-grasp-v.north west) + (-0.05, 0.13)$) -- ($(3-place-after.north east) + (0.05, 0.13)$);
    
    \node[above of=1-grasp-v, labels] (grasp-v-label) {Grasp Before \\ $s_{t;\, \textit{grasp}}$};
    \node[above of=1-place-v, labels] (place-v-label) {Place Before \\ $s_{t;\, \textit{place}}$};
    \node[above of=1-place-goal, labels] (place-goal-label) {Place Goal \\ $s_{t;\, \textit{goal}}$};
    \node[above of=1-place-after, labels] (place-after-label) {Place Result \\ $s_{t+1;\, \textit{place}}$};
\end{tikzpicture}
	
	\caption{Dataset samples from successful pick-and-place actions. Our approach uses four relevant observations, in particular the shown windows of the scene around the robot's \acf{TCP}. The robot grasps an object out of the \textit{grasp before} image, and places it into the \textit{place before} image with a given \textit{place goal} in mind. The subsequent \textit{place result} is only required for training. In a simplified manner, the robot should choose pick-and-place actions so that it cannot differentiate between the \textit{place goal} and \textit{place result} image.}
	\label{fig:data-sample}
\end{figure}
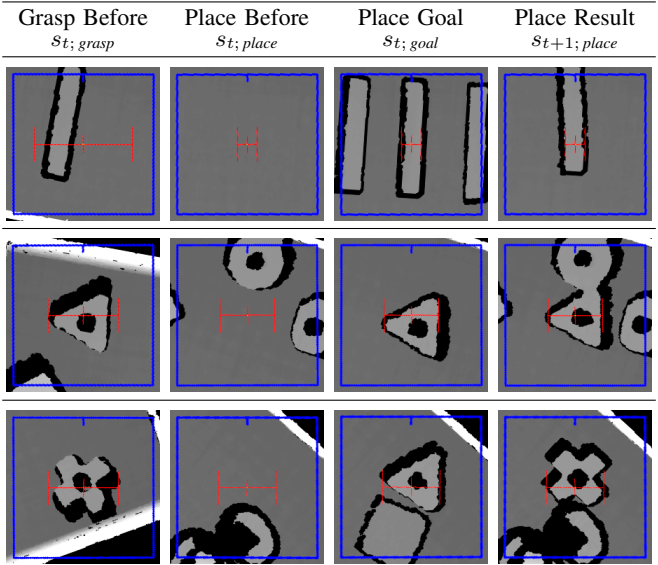

\begin{figure*}[t]
	\centering
	\vspace{0.8mm}
\begin{tikzpicture}[scale=0.8, framed]
    \node (image-mid) at (0, 0) {};
	\node[minimum width=4, text centered, above of=image-mid, node distance=115] (grasp-before-img) {Grasp Before};
	\draw[thick] (grasp-before-img.south) +(-1.5, -2.2) rectangle +(1.5, 0);
	
	\draw[red, fill=red] (grasp-before-img.south) +(0, -0.5) circle (0.2);
	\draw[green, fill=green] (grasp-before-img.south) ++(0.5, -0.5) -- +(0.2, 0.36) -- +(0.4, 0) -- cycle;
	\draw[blue, fill=blue, rotate=-24] (grasp-before-img.south) ++(0.05, -1.65) rectangle +(0.3, 0.3);
	\draw[rotate=-2] (grasp-before-img.south) ++(-0.8, -1.85) rectangle +(0.8, 0.8);

	\node[minimum width=4, text centered, above of=image-mid, node distance=35] (place-before-img) {Place Before};
	\draw[thick] (place-before-img.south) +(-1.5, -2.2) rectangle +(1.5, 0);
	
	\draw[yellow, fill=yellow] (place-before-img.south) +(-0.8, -0.8) circle (0.2);
	\draw[red, fill=red, rotate=-20] (place-before-img.south) ++(0.5, -1.6) -- +(0.2, 0.36) -- +(0.4, 0) -- cycle;
	\draw[rotate=24] (place-before-img.south) ++(-0.1, -1.2) rectangle +(0.8, 0.8);

	\node[below of=image-mid, node distance=45] (after-img) {};
	
	\node[minimum width=1, text width=74, text centered, left of=after-img, node distance=50] (place-after-img) {Measured Image Place Result};
	\draw[thick] (place-after-img.south) +(-1.5, -2.2) rectangle +(1.5, 0);
	
	\node[text width=30, align=center] (place-after-or) at ($(place-after-img.east) + (0.45, -1.6)$) {\textit{or}};
	
	\draw[yellow, fill=yellow] (place-after-img.south) +(-0.8, -0.8) circle (0.2);
	\draw[red, fill=red, rotate=-20] (place-after-img.south) ++(0.5, -1.6) -- +(0.2, 0.36) -- +(0.4, 0) -- cycle;
	\draw[blue, fill=blue, rotate=24] (place-after-img.south) ++(0.15, -0.92) rectangle +(0.3, 0.3);
	\draw[rotate=24] (place-after-img.south) ++(-0.1, -1.2) rectangle +(0.8, 0.8);
	
	\node[minimum width=1, minimum height=28, right of=after-img, node distance=50] (place-goal-img) {};
	\draw[thick] (place-goal-img.south) +(-1.5, -2.2) rectangle +(1.5, 0);
	
	\draw[yellow, fill=yellow] (place-goal-img.south) +(-0.8, -0.8) circle (0.2);
	\draw[red, fill=red, rotate=-20] (place-goal-img.south) ++(0.5, -1.6) -- +(0.2, 0.36) -- +(0.4, 0) -- cycle;
	\draw[blue, fill=blue, rotate=24] (place-goal-img.south) ++(0.0, -0.85) rectangle +(0.3, 0.3);
	
	\draw[rotate=24] (place-goal-img.south) ++(-0.1, -1.2) rectangle +(0.8, 0.8);

	\node[right of=image-mid, node distance=160] (mid-nn) {};
	\node[above of=mid-nn, node distance=50] (grasp-nn) {};
	\draw[] (grasp-nn) ++(-1.0, -1.5) rectangle +(0.25, 3);
	\draw[] (grasp-nn) ++(-0.6, -1.2) rectangle +(0.25, 2.4);
	\draw[] (grasp-nn) ++(-0.2, -0.9) rectangle +(0.25, 1.8);
	\draw[] (grasp-nn) ++(0.2, -0.7) rectangle +(0.25, 1.4);
    \draw[] (grasp-nn) ++(0.6, -0.6) rectangle +(0.25, 1.2);
	\node[draw, fill=white, thick, minimum width=55, minimum height=2.2em, text centered] (grasp-nn-label) at (grasp-nn) {Grasp NN};

	\node[below of=mid-nn, node distance=50] (place-nn) {};
	\draw[] (place-nn) ++(-1.0, -1.5) rectangle +(0.25, 3);
	\draw[] (place-nn) ++(-0.6, -1.2) rectangle +(0.25, 2.4);
	\draw[] (place-nn) ++(-0.2, -0.9) rectangle +(0.25, 1.8);
	\draw[] (place-nn) ++(0.2, -0.7) rectangle +(0.25, 1.4);
    \draw[] (place-nn) ++(0.6, -0.6) rectangle +(0.25, 1.2);
	\node[draw, fill=white, thick, minimum width=55, minimum height=2.2em, text centered] (place-nn-label) at (place-nn) {Place NN};
	
	\node[right of=mid-nn, node distance=80] (merge-minus) {};
    \draw[thick] (merge-minus) circle [radius=0.3] node {-};
	
	\node[right of=merge-minus, node distance=30] (merge-nn) {};
	\draw[] (merge-nn) ++(-0.3, -0.9) rectangle +(0.25, 1.8);
	\draw[] (merge-nn) ++(0.1, -0.5) rectangle +(0.25, 1.0);
    \draw[] (merge-nn) ++(0.5, -0.2) rectangle +(0.25, 0.4);

	\node[right of=merge-nn, node distance=36] (merge-reward-label) {};
	\fill (merge-reward-label) circle [radius=0.06];
	
	\node[text width=76, align=center] (merge-reward-label-top2) at ($(merge-reward-label) + (2.8, 1.1)$) {\textbf{Contrastive Loss:}};
	\node[text width=74, align=center] (merge-reward-label-top) at ($(merge-reward-label) + (2.8, 0.5)$) {measured image};
	\node[text width=74, align=center] (merge-reward-label-top3) at ($(merge-reward-label) + (2.8, 0.0)$) {\textit{or}};
	\node[text width=74, align=center] (merge-reward-label-bottom) at ($(merge-reward-label) + (2.8, -0.5)$) {negative sample};
	\node[text width=74, align=center] (grasp-reward-label) at ($(merge-reward-label) + (2.8, 3.2)$) {\textbf{Grasp Reward}};

	\draw[->, rounded corners=3pt] ($(grasp-nn-label.east) + (0, -0.2)$) to [out=345, in=105] ($(merge-minus.north) + (-0.04, 0.17)$);
	\draw[->, rounded corners=3pt] ($(place-nn-label.east) + (0, 0.2)$) to [out=15, in=255] ($(merge-minus.south) + (-0.04, -0.17)$);
	\draw[->, rounded corners=3pt] ($(merge-minus.east) + (0.16, 0)$) -- ($(merge-nn.west) + (-0.2, 0)$);
	
	\draw[->, rounded corners=3pt] ($(place-nn-label.east) + (0, -0.2)$) to [out=340, in=250] (merge-reward-label.south);
	
	\draw[->] ($(grasp-nn-label.east) + (0, 0.2)$) to [out=20, in=178] (grasp-reward-label.west);
    \draw[->] ($(merge-nn.east) + (0.6, 0)$) -- (merge-reward-label.west);
    
    \draw[->, dotted] ($(grasp-before-img.south) + (-0.4, -1.5)$) to [out=220, in=120] node[above, rotate=70]{\footnotesize \hspace{2em} pick-and-place object} ($(place-after-img.south) + (0.2, -0.2)$);
    \draw[->] ($(grasp-before-img.south) + (-0.2, -1.55)$) to [out=345, in=180] (grasp-nn-label.west);
    
    \draw[->] ($(place-before-img.south) + (0.85, -0.7)$) to [out=345, in=178] ($(place-nn-label.west) + (0, 0.2)$);
    \draw[->, dashed] ($(place-after-img.south) + (0.85, -0.5)$) to [out=40, in=180] ($(place-nn-label.west) + (0, -0.2)$);
    \draw[->, dashed] ($(place-goal-img.south) + (0.85, -0.5)$) to [out=40, in=180] ($(place-nn-label.west) + (0, -0.2)$);
    \node[minimum width=1, text width=70, text centered, right of=after-img, node distance=50, fill=white, fill opacity=0.7, text opacity=1] (place-goal-img2) {Negative Sample (e.g. Place Goal)};
    
    \draw[->, dashed] (merge-reward-label) -- (merge-reward-label-top.west);
    \draw[->, dashed] (merge-reward-label) -- (merge-reward-label-bottom.west);

    \node[] (z-g) at ($(grasp-nn-label.east) + (1.3, -0.8)$) {};
	\draw[fill=orange!50] (z-g) ++(-0.125, -0.8) rectangle +(0.25, 1.6);
	\node[draw, fill=white, thick, minimum width=20, minimum height=1.6em, text centered] (z-g-label) at (z-g) {$z_g$};
	
	\node[] (z-p) at ($(place-nn-label.east) + (1.3, 0.77)$) {};
	\draw[fill=orange!50] (z-p) ++(-0.125, -0.8) rectangle +(0.25, 1.6);
	\node[draw, fill=white, thick, minimum width=20, minimum height=1.6em, text centered] (z-p-label) at (z-p) {$z_p$};
\end{tikzpicture}

	\caption{Our \acf{NN} architecture during training: Given the images of both the grasping and placing scene, a system of three \acp{NN} predicts whether a given third image is a real measured image based on an executed action or a negative sample such as the goal image. While both the \textit{Grasp \ac{NN}} and the \textit{Place \ac{NN}} are limited to their corresponding scene, a third \ac{NN} predicts a combined contrastive loss via the difference between the $z_g$ and $z_p$ embeddings. We interpret this contrastive loss as a \textit{place reward}. Additionally, the fully-convolutional Grasp and Place \ac{NN} predict the grasp success and the place reward as an initial estimation, respectively.}
	\label{fig:nn-architecture}
\end{figure*}
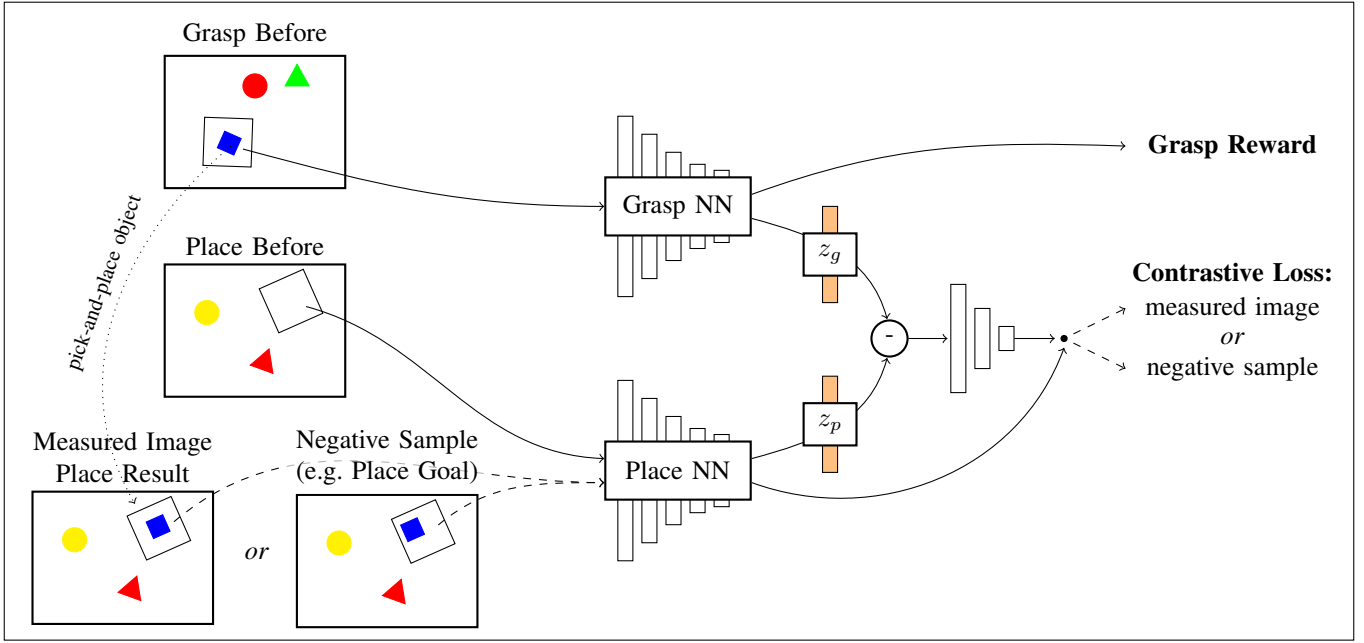

We define four \textit{grasp} primitives, which differ in the gripper opening as a pre-shaped gripper width. The robot approaches the given manipulation pose (or grasp point in this case) from above along the $z$-axis. If a collision is detected by its internal force sensors, the robot retracts a few millimeters. Then, the robot closes its fingers via force-control. A grasp success is measured if the closed gripper width is larger than zero in a given height above the bin. A single \textit{place} action opens the gripper at a given pose using the same approach trajectory.

While both learning to grasp from a dense reward and finding missing objects in the goal state is comparably easy, the challenge of precise pick-and-place is to find corresponding grasp and place actions. In this regard, we propose an approach to efficiently handle the Cartesian product $\mathcal{A}_g \times \mathcal{A}_p$ of both action spaces.

\subsection{Visual State Space}
\label{subsec:state-space}

Given the state space $\mathcal{S}$, let $s$ be a set of images in orthographic projection. Each image is either a depth or a RGBD image, depending on the chosen camera configuration. Due to the orthographic projection, each translation or rotation in the image space corresponds to a planar transformation $(x, y, \theta)$ of the robot in the task space. To observe the entire bin without occlusion, we use top-down views of the scene.

Let $s^\prime \subset s$ be an image window around the \acf{TCP} of the robot, which is defined by the tip of the closed gripper. As common in robot learning for manipulation, we train a \ac{FCNN} to estimate the action-value function $Q_a$ from an action given the corresponding image window $s^\prime$. During inference, we use the same \ac{FCNN} to estimate $Q_a$ at a grid of poses efficiently. This corresponds to a pixel-wise sliding-window approach for $(x, y)$, while $\theta$ is calculated by pre-rotating the image inputs. The window's side length is roughly equal to the maximal object size.
In terms of \ac{RL}, the policy $\pi(s) = \omega \circ Q_a(s, a)$ can be defined by a so-called selection function $\omega$ composed with the action-value function $Q_a$. We introduce different selection strategies $\omega$ for the training and inference phase in (\ref{subsec:inference}).

We consider the grasp and place actions to be in different scenes. Importantly, all images of the same scene are taken using the same camera pose. Then, we define the state space $\mathcal{S}$ by the set $s$ of following three observations:
\begin{enumerate}
    \item $s_{\texttt{grasp}}$, state of the grasp scene before the grasp $a_g$,
    \item $s_{\texttt{place}}$, state of the place scene before the place $a_p$, and
    \item $s_{\texttt{goal}}$, goal state of the place scene. Usually, this is an image of a scene configured by a human demonstrator.
\end{enumerate}

Furthermore, the state of the place scene after the place $a_p$ is a fourth important observation. It is denoted as $s_{t+1;\, \texttt{place}}$ and can be understood as the \textit{result} of a pick-and-place action. It is only available and required for training. Examples of the four relevant observations are shown in Fig.~\ref{fig:data-sample}.

\subsection{Learning from Goal States}
\label{subsec:reward-function}

Since the robot should imitate the given goal state, the reward $r$ needs to capture the similarity between $s_{\texttt{goal}}$ and the expected $s_{t+1;\, \texttt{place}}$, conditioned on both before states $s_{\texttt{grasp}}$ and $s_{\texttt{place}}$. In general, the robot should choose those actions that most probably result in the \textit{place goal} image as the real \textit{place result} image. We interpret this as a task of \textit{density estimation} and apply \ac{NCE} \cite{gutmann2010noise} to this unsupervised learning problem. Given a dataset $D$ of pick-and-place actions $a$, a model is trained to discriminate between the real data $D$ (with label $C=1$) and a noise distribution $Q$ (with $C=0$). Then, a classifier is able to measure the probability $\xi$ of an image $s^\prime$ being a realistic outcome of a pick-and-place action:
\begin{equation}
    \xi(s^\prime) := p(C=1 | s^\prime, s^\prime_{\texttt{grasp}}, s^\prime_{\texttt{place}}) \nonumber
\end{equation}
We refer to this probability as the \textit{place reward} $\xi \in [0, 1]$. Besides, the estimated \textit{grasp reward} $\psi \in [0, 1]$ of the binary grasp success can be interpreted like a grasp probability. 

For \ac{NCE}, the design of the noise distribution $Q$ is crucial for estimating the density of the data $D$. For each sample in $D$ with a given grasping image $s^\prime_{\texttt{grasp}}$, we augment the remaining pair of place images. For the data distribution ($C=1$) in a pick-and-place task, we generate two positive samples:

\begin{LaTeXdescription}
    \item[Hindsight] Foremost, the real measured pair of $\left( s^\prime_{\texttt{place}}, s^\prime_{t+1;\, \texttt{place}} \right)$ is the basic positive sample in the data distribution.
    \item[Further Hindsight] If further objects are placed into the same bin for $t_{\texttt{bin}}$ steps after the current pick-and-place action, the images $\left( s^\prime_{\texttt{place}}, s^\prime_{t+n;\, \texttt{place}} \right)$ for $1 \leq n \leq t_{\texttt{bin}}$ are used as positive samples as well. This sample enables the robot to consider future actions when multiple pick-and-place actions are needed to achieve a goal state, e.g.\ for box stacking.
\end{LaTeXdescription}
For the noise distribution ($C=0$), we generate a range of negative samples as follows:
\begin{LaTeXdescription}
    \item[Negative Foresight] We create negative pairs using either identical images (for both $s^\prime_{\texttt{place}}$ and $s^\prime_{t+1;\, \texttt{place}}$) or positive pairs in the wrong temporal order $\left( s^\prime_{t+1;\, \texttt{place}}, s^\prime_{\texttt{place}} \right)$.
    \item[Augmented Hindsight] Moreover, we generate additional negative samples by jiggling the pose of the place images $s^\prime$. In particular, we jiggle real hindsight images with a minimum displacement of \SI{1}{mm} or \SI{3}{{}^\circ}.
    \item[Goal] The goal image pair $\left( s^\prime_{\texttt{place}}, s^\prime_{\texttt{goal}} \right)$ is used as a negative sample. As the training progresses and accuracy improves, both images should converge and lead to a median contrastive loss. To circumvent this effect, we apply methods of \textit{confident learning}: If a trained classifier predicts a goal image sample to be from the real data distribution with given certainty, we remove the goal image from the noise distribution furthermore.
    \item[Other Hindsight] Finally, place images of independent actions $a_g$ and $a_p$ are used. In particular, this results in mismatching object types as negative samples.
\end{LaTeXdescription}
Using \ac{NCE}, the imitation learning problem simplifies to ordinary supervised learning. The contrastive loss is defined using the \ac{BCE}
\setlength{\arraycolsep}{0.0em}
\begin{eqnarray}
    y_i &{}&{}= z_i - \log p(i) \nonumber \\
    \text{BCE} &&{}= \sum_i^N C_i \log \sigma(y_i) - (1 - C_i) \log (1 - \sigma(y_i))
    \label{eq:bce}
\end{eqnarray}
\setlength{\arraycolsep}{5pt}
with the logits $z_i$ adapted by the probability $p$ of the sample $i$, and the sigmoid function $\sigma$. We train a system of three \acp{NN} by minimizing the \ac{BCE} of the contrastive loss (eq.~\ref{eq:bce}) as well as the binary grasp reward. However, three constraints are applied to our \ac{NN} architecture (Fig.~\ref{fig:nn-architecture}): First, we define two separate Grasp and Place \acp{NN} as \acp{FCNN} and limit their input to their corresponding scene. Second, the Grasp \ac{NN} predicts the grasp reward as an additional output. Similarly, the Place \ac{NN} pre-estimates the place reward, however without information about the grasp action. We denote this prediction by the Place \ac{NN} as $\xi_p$. Third, both \acp{NN} calculate action embeddings $z_g$ or $z_p$ respectively. They are combined in the (non-convolutional) Merge \ac{NN} using their element-wise difference. Since information from both the grasp and the place scene are joined here, the place reward $\xi$ can then be predicted with significantly improved accuracy.

\subsection{Pick-and-place Inference}
\label{subsec:inference}

\begin{figure}[t]
	\centering
	\subfloat[Grasp reward $\psi$]{
		\includegraphics[trim=90 47 60 50, clip, height=0.48\linewidth, angle=90, origin=c]{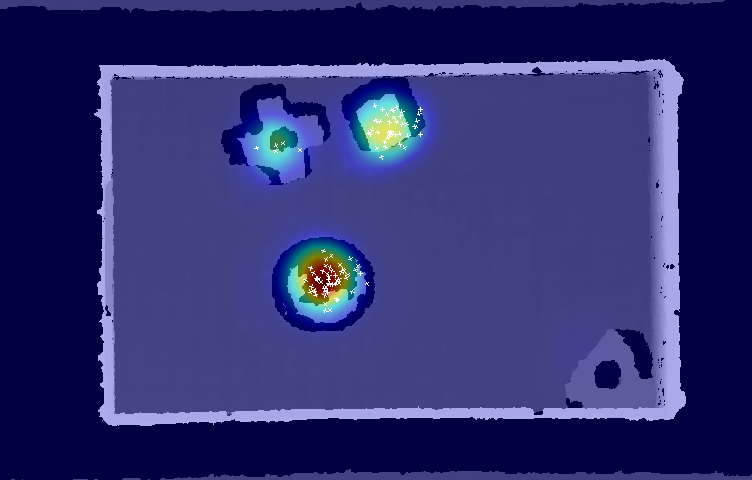}
	}
	\subfloat[Place reward $\xi_p$]{
		\includegraphics[trim=93 47 57 50, clip, height=0.48\linewidth, angle=90, origin=c]{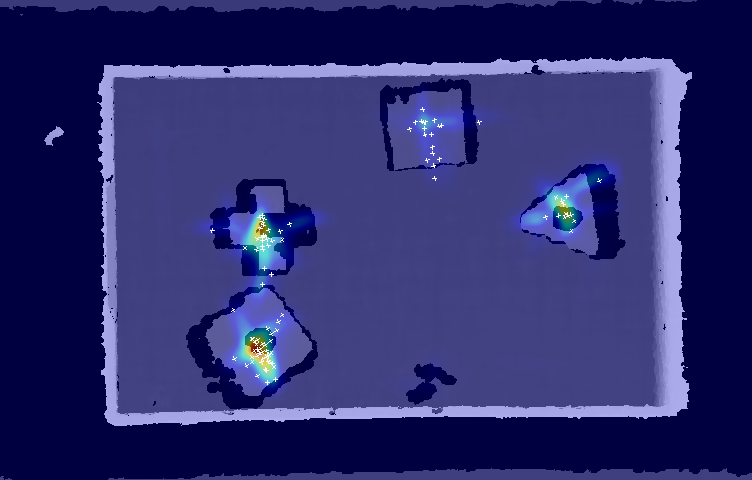}
	}
	
	\caption{Example heatmaps of the estimated grasp and place reward, ranging from low (blue) to high (red). The estimations of each \ac{FCNN} are averaged over rotations. For visualization, we reduce the number of sampled actions proposals to $N=80$ (white dots).}
	\label{fig:heatmaps}
\end{figure}

During the inference phase, we first rotate the images $s$ of the grasp and place scene given a discrete set of rotations (Fig.~\ref{algo:inference}). The image batches are fed into their corresponding \ac{FCNN}, resulting in action rewards $\psi$ and $\xi_p$ with their corresponding embeddings $z_g$ and $z_p$ for a discrete set of action poses. The final action space $\mathcal{A}$ for pick-and-place scales by the number of combinations $\mathcal{A}_g \times \mathcal{A}_p$. For performance reasons, not all combinations can be evaluated in the Merge \ac{NN}. Therefore, a small fraction of grasp and place actions are pre-selected using the probability distribution
\begin{equation}
    p(a_g | \psi) = \frac{\psi^\alpha}{\sum_{\mathcal{A}_g} \psi^\alpha}, \quad p(a_p | \xi_p) = \frac{\xi_p^\alpha}{\sum_{\mathcal{A}_p} \xi_p^\alpha}, \quad \alpha > 1. \nonumber
\end{equation}
Furthermore, we set $\alpha = 6$ and sample $N := N_g = N_p = 200$ action proposals without replacement, respectively. Fig.~\ref{fig:heatmaps} illustrates the grasp and place reward $\psi$ and $\xi_p$ as heatmaps, moreover indicating the position of sampled action proposals.
\begin{figure}[h]
    \begin{algorithmic}[1]
        \Data Images $s_{\texttt{grasp}}$, $s_{\texttt{place}}$, $s_{\texttt{goal}}$
        \Result Grasp $a_g$, Place $a_p$
        \Statex
        
        \State Create set of rotated grasp images $S_g$  
        \State Create set of rotated place and goal images $S_p$
        \State Calculate grasp rewards $\psi$ and embeddings $z_g$ for each pose in $S_g$ using Grasp \ac{FCNN}
        \State Calculate place rewards $\xi_p$ and embeddings $z_p$ for each pose in $S_p$ using Place \ac{FCNN}
        \State Sample $N_g$ grasps $z_{gi}$ with probability $(\psi_i)^\alpha$
        \State Sample $N_p$ places $z_{pj}$ with probability $(\xi_{p;j})^\alpha$
        
        \For{each combination $(z_{gi}, z_{pj})$}
            \State Subtract $z_{gi}$ and $z_{pj}$ element-wise
            \State Calculate final place reward $\xi$ using Merge NN
        \EndFor

        \State Select actions greedily via $a_g, a_p = \argmax_{ij} \psi \cdot \xi$
    \end{algorithmic}
    
    \caption{Algorithm of inferring a pick-and-place action using \acp{FCNN} and reward pre-estimation.}
    \label{algo:inference}
\end{figure}
For each combination of proposed grasp $a_g$ and place $a_p$, the Merge \ac{NN} predicts the refined place reward $\xi$ from $z_g - z_p$. Since we only train the Place and Merge \ac{NN} on pick-and-place actions with successful grasps, the place reward $\xi$ is conditioned on $\psi=1$. This way, we extend common approaches for learning \textit{grasping}, and are able to learn the grasp confidence with additional grasp data independently. Finally, the pick-and-place reward $r=\psi \cdot \xi$ is defined by the product of grasp and place reward.

One of three selection functions $\omega$ is applied to the set of $N^2$ proposed pick-and-place actions: First, a uniform \textit{random} selection for initial exploration. Second, a \textit{sampling} based approach using $a_g, a_p \sim \psi \cdot \xi$. Third and most important, a \textit{greedy} selection method $a_g, a_p = \argmax \psi \cdot \xi$ is used for application. Then, given the scene images, the robot chooses the grasp and place combination that makes the scene after the executed action look most like the goal image.

\subsection{Self-supervised Learning}

In order to scale the real-world training, the learning process needs to work with minimal human interaction. Therefore, two bins are used for continuous and diverse training. We apply a simple curriculum learning strategy, starting with single objects and increasing the number and complexity of object types over time.
In the beginning, we sample grasps from a given grasping policy and place objects in the second bin randomly. Later on, we sample goal states from a range of previously seen before states, denoted as the goal database.
As our approach is off-policy, we train the \ac{NN} using the most current dataset in parallel to the real-world data collection. Then, we sample the pick-and-place action with an $\varepsilon$-\textit{sample} strategy, reducing $\varepsilon$ over time. In the end of the training, we switch to the $\varepsilon$-\textit{greedy} approach. 

Optionally, the training of the grasp reward can be improved using single grasps without place action. As pick-and-place is an extension to more common bin picking pipelines, the training can then be bootstrapped by reusing prior data.

\section{Experimental Results}

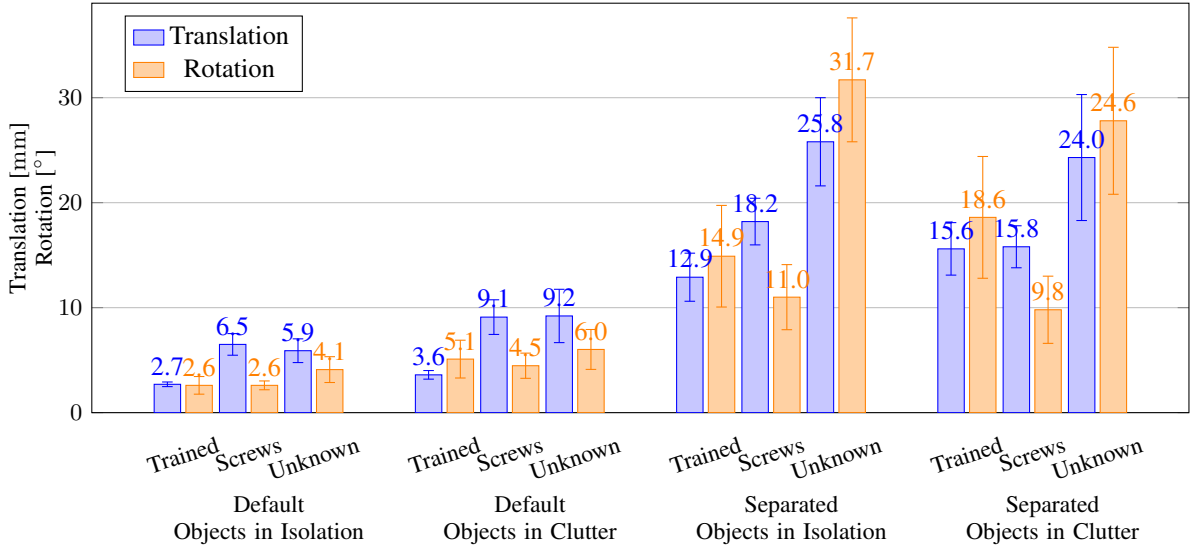
\begin{figure*}[t]
    \centering
\begin{tikzpicture}
\begin{axis}[
  xticklabels={Trained, Screws, Unknown, Trained, Screws, Unknown, Trained, Screws, Unknown, Trained, Screws, Unknown},
  xtick=data,
  ylabel={Translation [\si{mm}] \\ Rotation [\si{^\circ}]},
  ylabel style={text width=80, align=center, yshift=-6pt},
  major x tick style=transparent,
  ybar,
  ymin=0,
  ymax=39,
  width=460,
  height=200,
  xtick=data,
  ymajorgrids=true,
  nodes near coords,
  point meta=explicit symbolic,
  legend pos=north west,
  legend image code/.code={%
      \draw[#1] (0cm, -0.1cm) rectangle (0.4cm, 0.1cm);
  },
  extra x ticks={2, 6, 10, 14},
  extra x tick labels={
    {Default \\ Objects in Isolation},
    {Default \\ Objects in Clutter},
    {Separated \\ Objects in Isolation},
    {Separated \\ Objects in Clutter},
},
  label style={font=\small},
  ticklabel style={font=\small},
  x tick label style={rotate=18},
  every extra x tick/.style={
        xticklabel style={
            yshift=-25pt,
            rotate=-18,
            text width=80,
            align=center,
        },
    },
]

\addplot[color=blue, fill=blue!22, error bars/.cd, y dir=both, y explicit] coordinates {
    (1, 2.7) +- (0, 0.22)[2.7]
    (2, 6.5) +- (0, 1.03)[6.5]
    (3, 5.9) +- (0, 1.13)[5.9]
    
    (5, 3.6) +- (0, 0.41)[3.6]
    (6, 9.1) +- (0, 1.65)[9.1]
    (7, 9.21) +- (0, 2.54)[9.2]
    
    (9, 12.9) +- (0, 2.29)[12.9]
    (10, 18.2) +- (0, 2.22)[18.2]
    (11, 25.8) +- (0, 4.2)[25.8]
    
    (13, 15.6) +- (0, 2.5)[15.6]
    (14, 15.8) +- (0, 2.0)[15.8]
    (15, 24.3) +- (0, 6.0)[24.0]
};

\addplot[color=orange, fill=orange!36, error bars/.cd, y dir=both, y explicit] coordinates {
    (1, 2.6) +- (0, 0.84)[2.6]
    (2, 2.6) +- (0, 0.42)[2.6]
    (3, 4.1) +- (0, 1.23)[4.1]
    
    (5, 5.1) +- (0, 1.8)[5.1]
    (6, 4.47) +- (0, 1.2)[4.5]
    (7, 6.02) +- (0, 1.9)[6.0]
    
    (9, 14.9) +- (0, 4.84)[14.9]
    (10, 11.0) +- (0, 3.1)[11.0]
    (11, 31.7) +- (0, 5.9)[31.7]
    
    (13, 18.6) +- (0, 5.8)[18.6]
    (14, 9.8) +- (0, 3.2)[9.8]
    (15, 27.8) +- (0, 7.0)[24.6]
};

\legend{Translation, Rotation}
\end{axis}
\end{tikzpicture}
    \caption{The translational and rotational mean placement error in different settings. We compare the \textit{default} case ($N$=200) with the \textit{separated} case where the best grasp and best place actions are chosen independently ($N$=1). Here, $N$ is the number of proposed action embeddings for further combination. Moreover, we differentiate between experimental results for grasping isolated objects and out of clutter.}
    \label{fig:placement-error-barchart}
\end{figure*}

For our real-world experiments, a Franka Emika Panda robot including the default gripper with custom made jaws (Fig.~\ref{fig:overall-system}) were used. Both an Ensenso N10 depth- and a RealSense D435 RGBD-camera are mounted on the flange. The system uses an Intel Core i7-8700K processor and a NVIDIA GTX 1070 Ti for computing. In front of the robot, two bins with a variety of objects for interaction are placed during training.

We crop and scale the camera images to $32 \times 32$ pixels during training and $110 \times 110$ during inference, resulting in an effective translational resolution of around \SI{3}{mm} for place and \SI{6}{mm} for grasp actions. However, the final pick-and-place precision may fall below this value if matching grasp and place actions are inferred. We use \num{37} image rotations, leading to an overall action space size of both \num{236800} grasps (enlarged by four grasping primitives) and places, as well as \num{5.6e10} pick-and-place actions. The embedding size of $z_g$ and $z_p$ is set to \num{48}. Calculating the next action takes around \SI{400}{ms}. \\

Further details, the source code, more dataset samples, and supplementary videos showing our experimental results are published at \textit{\url{https://pantor.github.io/learning-pick-and-place/}}.

\subsection{Data Collection and Training}

We evaluate our approach with two distinct models: First, a \textit{specialized} model for pick-and-placing screws (M10$\times$60) using RGBD images. This model was trained with around \num{3500} pick-and-place actions. We reuse \num{12000} sole grasps from prior experiments for improving the grasp reward estimation of the screw model. Second, a \textit{general} model was trained for all remaining object types and experiments. Due to reliability issues of the RealSense camera, this model uses only depth-images of the Ensenso N10. It was trained on wooden primitive shapes with side lengths of $\approx$\,\SI{4}{cm} (Fig.~\ref{fig:overall-system}) for around \num{25000} pick-and-place actions, corresponding to around \SI{120}{h}.

Both Grasp and Place \acp{NN} are fully-convolutional and share the first few layers between the reward and embedding outputs, respectively. The merge \ac{NN} is a three-layer dense neural network. We double the loss weight of the final place reward and optimize the \acp{NN} using Adam with a learning rate of \num{2e-4}. After \num{100} epochs, we remove goal images with a predicted contrastive loss of above \num{0.7} from the training set. For further details, we refer to our open-sourced implementation linked above.

\subsection{Object Placement Error}

The precision of the pick-and-place task is evaluated using the placement error of a single object. We define this error as the distance between the object's pose within the goal state and the pose after the executed pick-and-place action. Given a high repetition accuracy of the robot and a well-calibrated depth-camera, the goal and result images are taken from the same camera pose. Then, we measure the placement error by determining the 2D transformation bringing $s^\prime_{\texttt{goal}}$ and $s^\prime_{t+1;\, \texttt{place}}$ in alignment.

Fig.~\ref{fig:placement-error-barchart} shows the translational and rotational placement error in various settings. The placement error is investigated for isolated objects as well as objects in clutter. For the latter, we fill the grasping bin with 25 trained objects, 80 screws, or 10 unknown objects respectively and measure only places of the correct object type. Additionally, we compare the results of our proposed default system with a separated approach: 
\begin{figure}[b]
  \centering
  \includegraphics[trim=0 15 0 50, clip, width=0.82\linewidth]{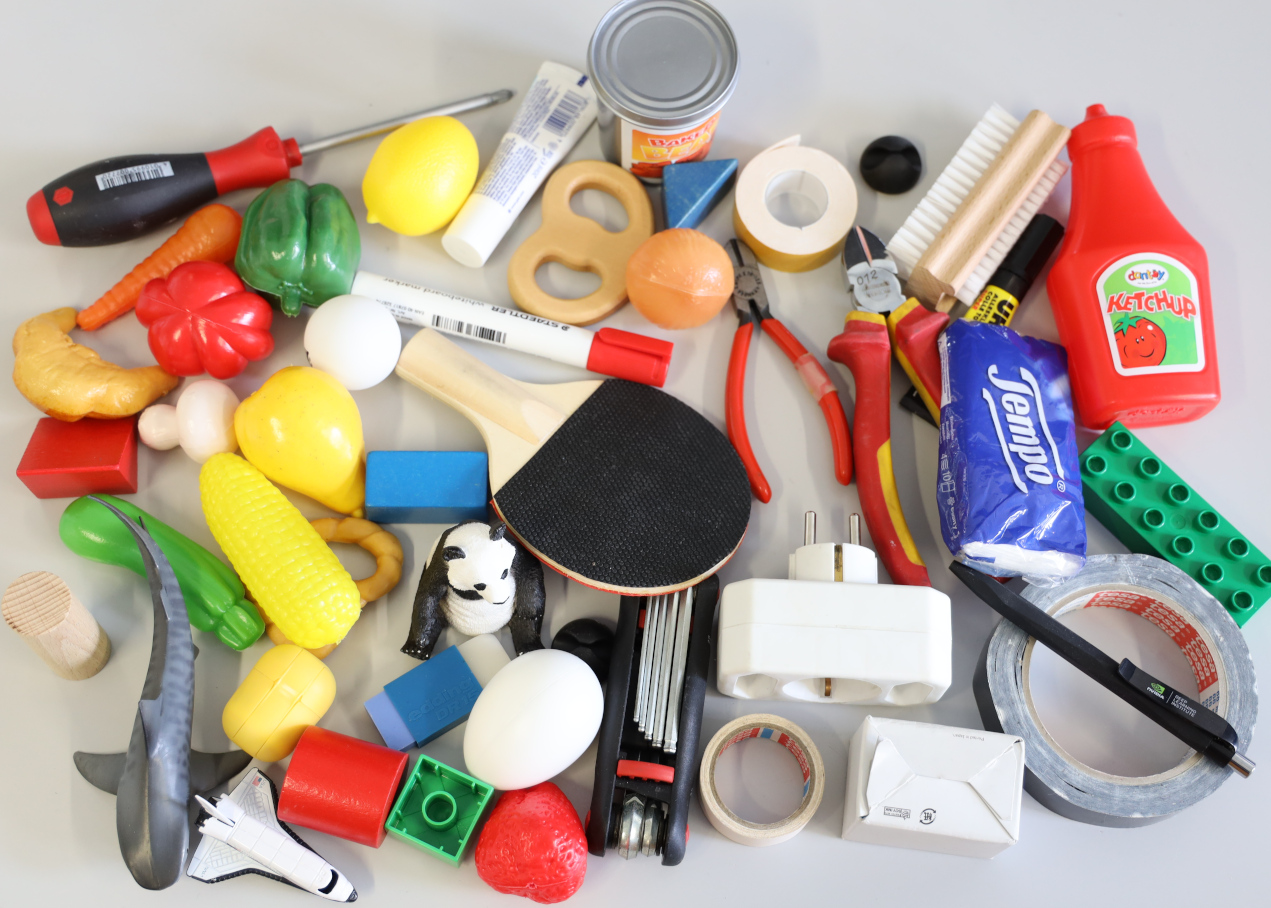}
  \caption{Our evaluation set of \num{50} unknown objects.} 
  \label{fig:unknown-objects}
\end{figure}
Then, we set $N=1$, leading to a system that chooses the best grasp action and the best place action independently of each other, and does not make use of the Merge \ac{NN}.

For trained objects, our robot achieves an average placement error below of \SI{3}{mm} and \SI{3}{^\circ}. Interestingly, the translational error falls just below the placing resolution. We assume the worse results for screws to be caused by less training data and a more complex, e.g.\ reflective, visual appearance. For unknown objects in our test set (Fig.~\ref{fig:unknown-objects}), the robot is still able to achieve an average precision of around \SI{6}{mm} and \SI{4}{^\circ}. In clutter, the precision decreases in particular for screws and unknown objects. Here, typical grasp rates lie around \SI{95}{\%} for trained objects, and around \SI{85}{\%} for unknown objects. Over all experiments, we find an average error of \SI{1.6}{mm} in the direction of the gripper jaws constraining the object, and \SI{3.9}{mm} orthogonal thereto. These findings suggest that the initial object displacement caused by the closing gripper might influence the placing precision significantly.

\subsection{Insertion Task}

Although the robot did not learn to insert objects directly, we investigate this capability despite tolerances of around \SI{1}{mm} using the peg game (Fig.~\ref{fig:overall-system}), pushing the robot to its pick-and-place precision limits. On average, the robot achieves a success rate of \SI{72}{\%} for the insertion of isolated objects, however depending heavily on the object type (Table~\ref{tab:results-insertion}).
\begin{table}[hb]
	\centering
	\renewcommand{\arraystretch}{1.15}  
	\caption{Success rates of inserting a given object onto a peg}
	\label{tab:results-insertion}
	\begin{tabular}{c|c|c|c|c}
	\hline
	& \multicolumn{2}{c|}{Default} & \multicolumn{2}{c}{Separated} \\
    Object & Isolation & Clutter & Isolation & Clutter \\ 
	\hline
	\hline
	Circle & \num{9} / \num{10} & \num{7} / \num{10} & \num{5} / \num{10} & \num{1} / \num{10} \\
	Triangle & \num{8} / \num{10} & \num{4} / \num{10} & \num{0} / \num{10} & \num{0} / \num{10} \\
	Square & \num{9} / \num{10} & \num{7} / \num{10} & \num{1} / \num{10} & \num{0} / \num{10} \\
	Oval & \num{6} / \num{10} & \num{6} / \num{10} & \num{3} / \num{10} & \num{0} / \num{10} \\
	Cross & \num{4} / \num{10} & \num{2} / \num{10} & \num{1} / \num{10} & \num{0} / \num{10} \\
	\hline
	& \SI{72}{\%} & \SI{52}{\%} & \SI{20}{\%} & \SI{2}{\%} \\
	\hline
	\end{tabular}
\end{table}
For example, we observed that predicting the displacement of the cross shape during clamping is difficult to do - again suggesting that the grasp displacement is a primary source of imprecision. The system is able to increase success rates by around \SI{50}{\%} in comparison to a separated approach. Moreover, the peg game also represents a perturbation of the environment. Success rates of up to \SI{90}{\%} suggest that the system is able to generalize to unknown environments. We classify a wrong object type while grasping in clutter as a failure.

\subsection{Selection Task}

\begin{figure}[ht]
	\centering
\begin{tikzpicture}[
	description/.style={align=left, text width=8cm, font=\small}
]
\begin{axis}[
	xbar=5pt,
	xmajorgrids,
	bar width=12pt, 
	width=0.52\textwidth,
	height=4.1cm,
	xmin=0,
	xmax=100,
	ymin=-1,
	ymax=1,
	nodes near coords,
    nodes near coords align=horizontal,
    nodes near coords style={anchor=west},
	yticklabels={,,},
	xlabel={Success Rate [\%]},
]

\addplot[color=black!80, fill=black!24, error bars/.cd, x dir=both, x explicit] coordinates {(20,0)[20]};
\addplot[color=black!80, fill=black!6, error bars/.cd, x dir=both, x explicit] coordinates {(74,0) +- (6, 0)[70]};
\addplot[color=black!60!orange, fill=orange!36, error bars/.cd, x dir=both, x explicit] coordinates {(60,0) +- (7, 0)[60]};
\addplot[color=blue, fill=blue!20, error bars/.cd, x dir=both, x explicit] coordinates {(86,0) +- (5, 0)[86]};

\end{axis}

\node[description, color=black!10!blue] at (4.1, 2.13) {Trained Objects};
\node[description, color=black!60!orange] at (4.1, 1.53) {Unknown Objects};
\node[description, color=black!90] at (4.1, 0.93) {Template Matching};
\node[description, color=black!96] at (4.1, 0.36) {Random};
\end{tikzpicture}
	\caption{Success rates of grasping the demonstrated objects out of a set of five distinct objects (\num{1}-out-of-{5} selection task), independent of the final place precision. This comparison uses depth images only.}
	\label{fig:results-selection}
\end{figure}
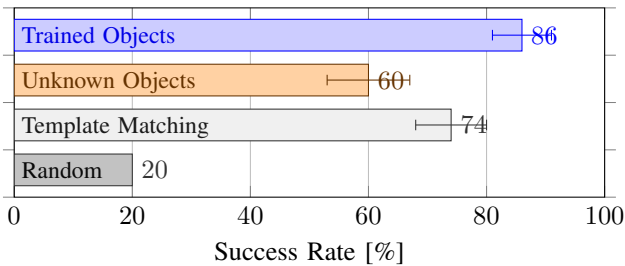

Given multiple object types in the grasping scene, the task of selecting the correct object to place arises naturally. We evaluate this task by choosing five objects randomly from the set of either all known or unknown (Fig.~\ref{fig:unknown-objects}) objects. Then, the robot needs to select the solely shown object from the goal state. For this \num{1}-out-of-\num{5} selection task, the robot achieves a success rate of \SI{86 \pm 5}{\%} for trained, and \SI{60 \pm 7}{\%} for unknown objects, in comparison to a random success rate of \SI{20}{\%} (Fig.~\ref{fig:results-selection}).
Note however, that the evaluated model does not make use of color images, which we assume to be very helpful for this task. We additionally compare our model against a simple \textit{template matching} baseline. First, the target object is detected using the difference between the goal and place image. Second, we apply this template to match a corresponding object within the grasp image. Although we find that this baseline is sensitive to depth shadows, it still outperforms our approach for unknown objects.

\subsection{Multiple-step Tasks}

Our approach does also allow to infer multiple actions from a single goal image, while updating the grasp and place images after each action. Given a single demonstration, the robot is able to place multiple objects out of clutter, with examples shown in (Fig.~\ref{fig:examples-multiple-steps-kit} and~\ref{fig:examples-multiple-steps-screws}). Moreover, we can take a sequence of goal images as an instruction list. This allows a wide range of easy-programmed pick-and-place tasks (Fig.~\ref{fig:examples-multiple-steps-building}). Videos of all three examples are included in our supplementary material linked above.

\tikzset{
  double arrow/.style args={#1 colored by #2 and #3}{
    -stealth,line width=#1,#2, 
    postaction={draw,-stealth,#3,line width=(#1)/3, shorten <=(#1)/3,shorten >=2*(#1)/3}, 
  }
}

\begin{figure}[ht]
	\centering
    \vspace{0.8mm}

	\subfloat[Placing the logo of our Alma Mater (KIT) (6 actions, 1 goal).]{
	\begin{tikzpicture}
        \node[anchor=east,inner sep=0] at (0,0) {\includegraphics[trim=180 190 60 100, clip, width=0.48\linewidth]{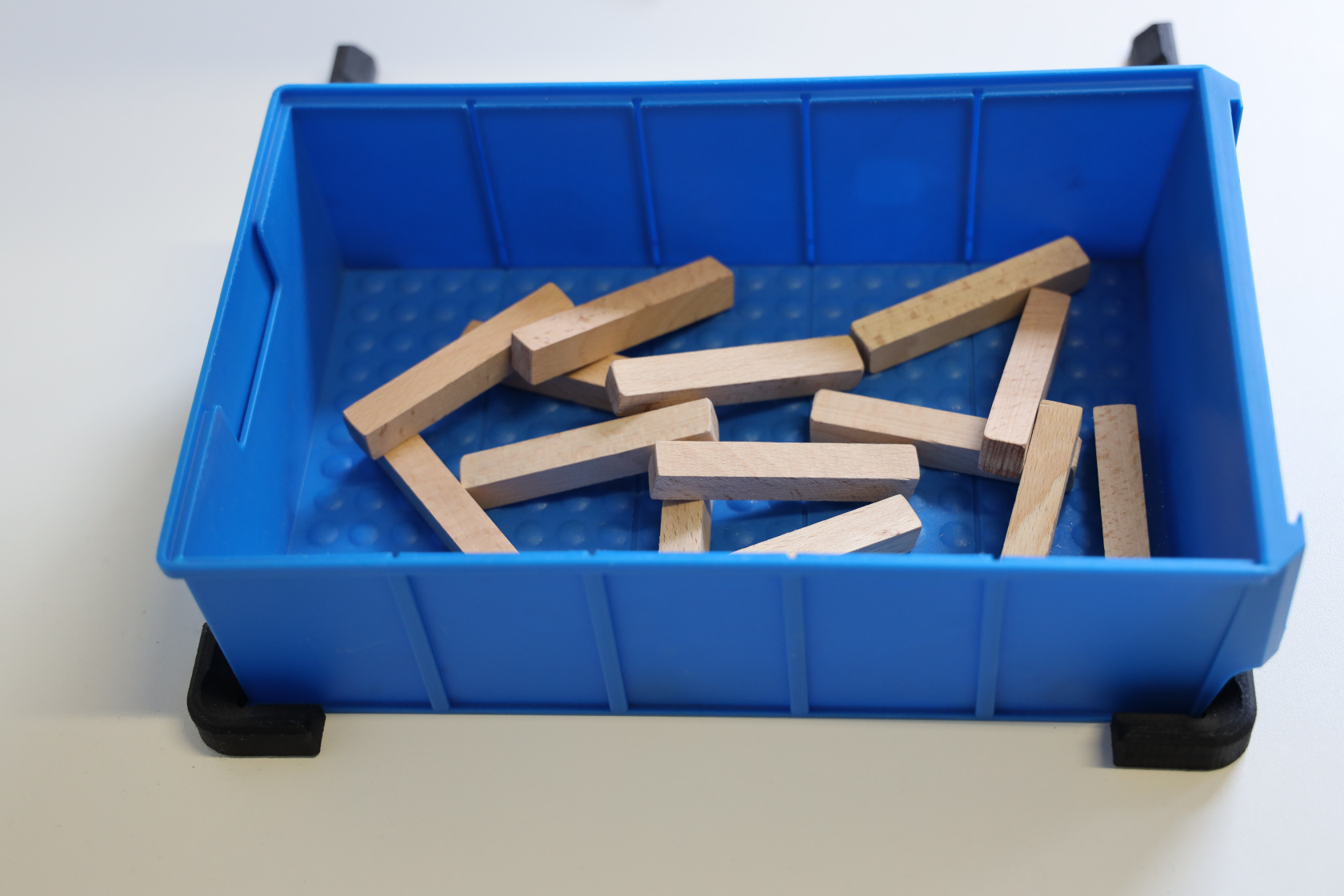}};
        \node[anchor=west,inner sep=0] at (0.1,0) {\includegraphics[trim=180 190 60 100, clip, width=0.48\linewidth]{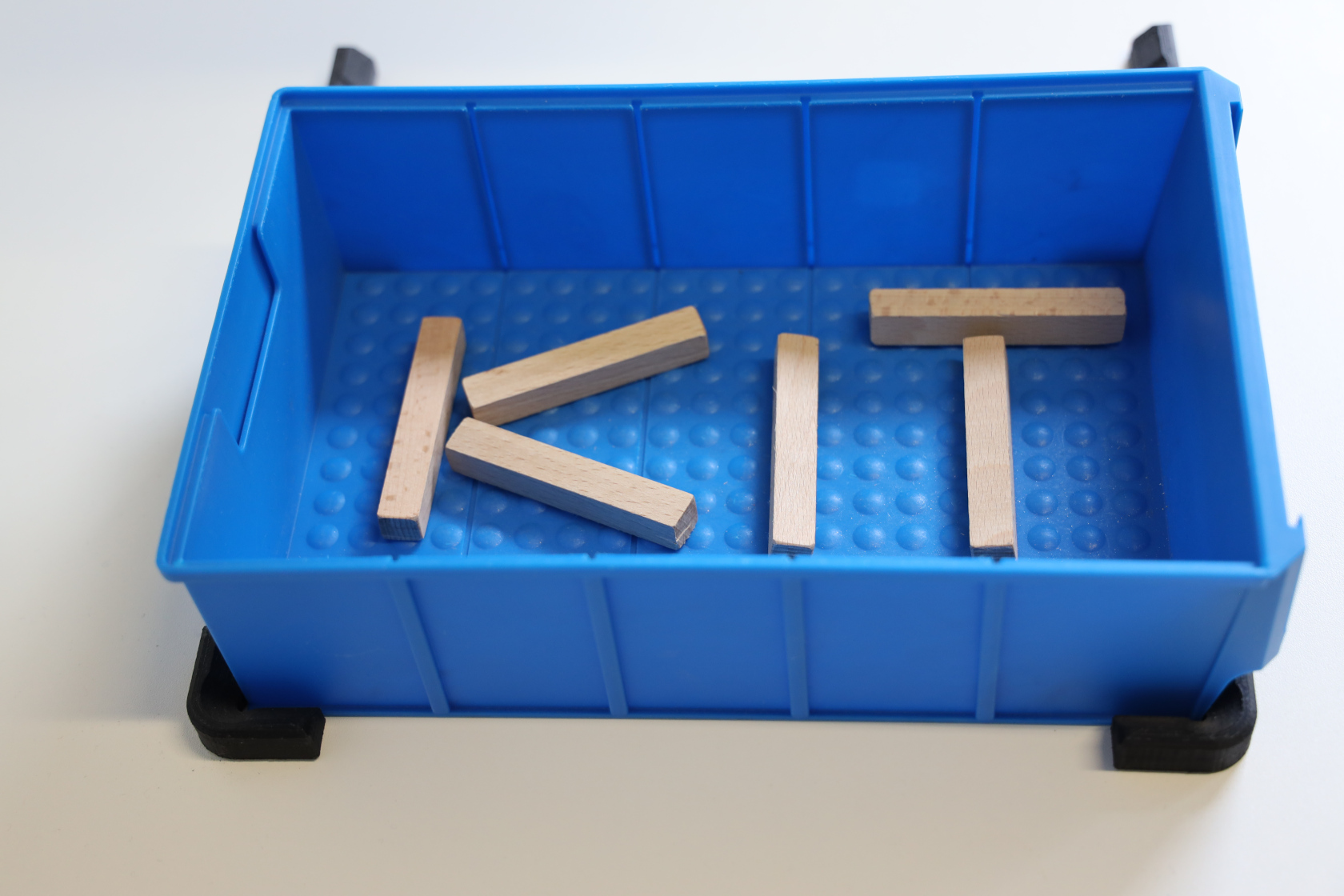}};
        
        \draw[double arrow=6pt colored by black and white] (-0.6,0.2) to [out=30, in=150] (0.7,0.2);
    \end{tikzpicture}
		\label{fig:examples-multiple-steps-kit}
	}
	
	\subfloat[Placing screws in an industrial scenario (3 actions, 1 goal).]{
	\begin{tikzpicture}
        \node[anchor=east,inner sep=0] at (0,0) {\includegraphics[trim=180 190 60 100, clip, width=0.48\linewidth]{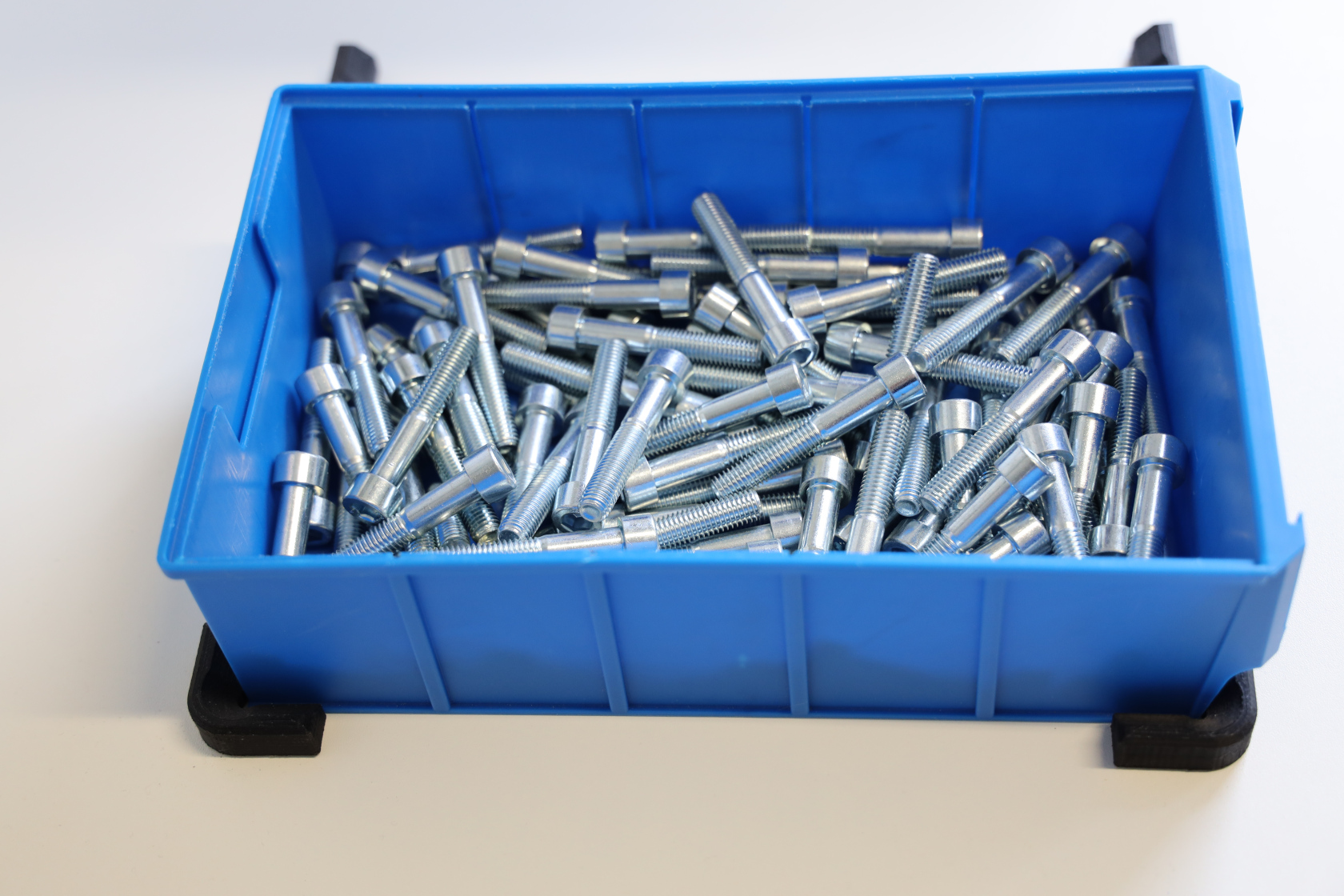}};
        \node[anchor=west,inner sep=0] at (0.1,0) {\includegraphics[trim=180 190 60 100, clip, width=0.48\linewidth]{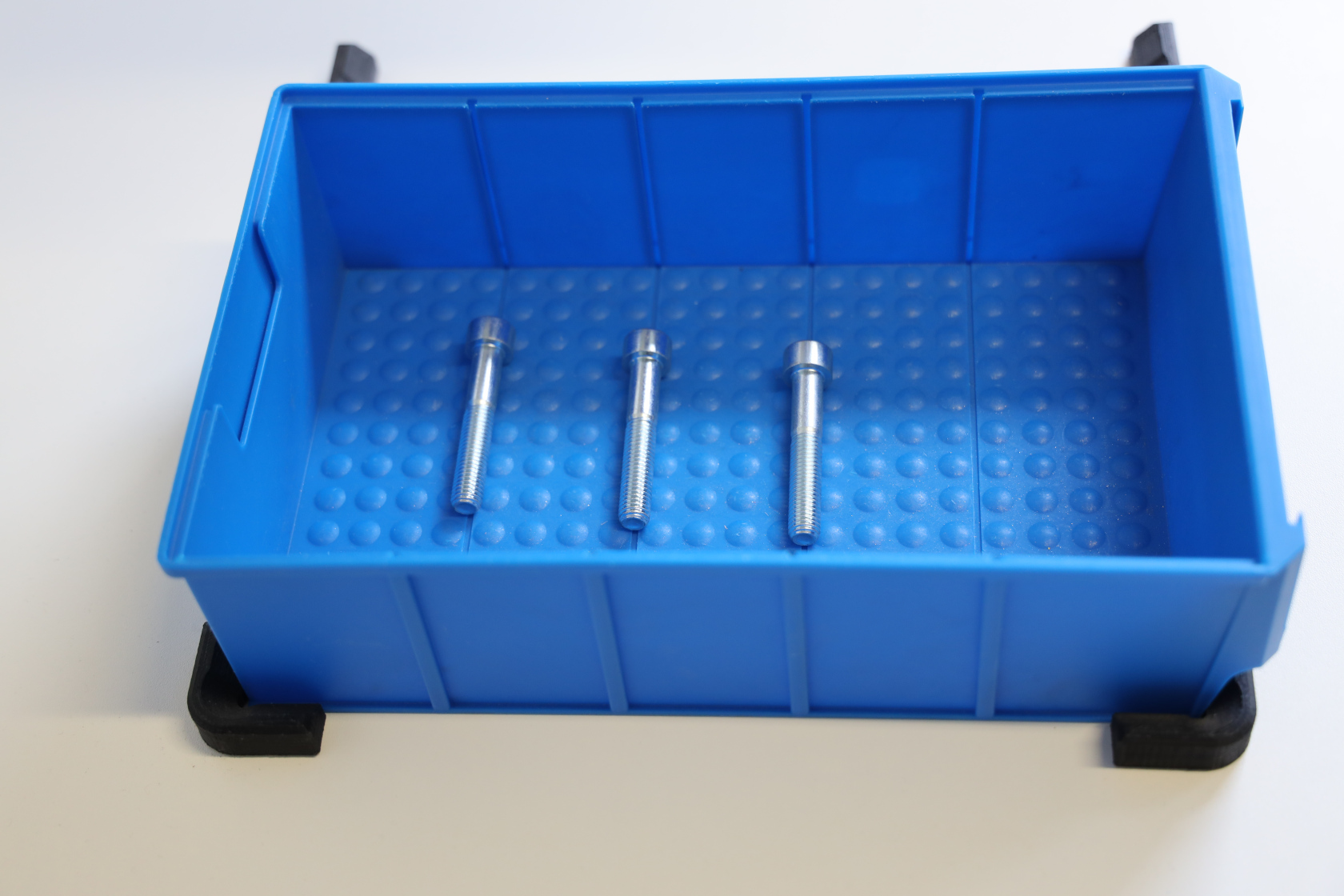}};
        
        \draw[double arrow=6pt colored by black and white] (-0.6,0.2) to [out=30, in=150] (0.7,0.2);
    \end{tikzpicture}
		\label{fig:examples-multiple-steps-screws}
	}
	
	\subfloat[Building a house with wooden blocks (6 actions, 4 goals).]{
	\begin{tikzpicture}
        \node[anchor=east,inner sep=0] at (0,0) {\includegraphics[trim=180 190 60 100, clip, width=0.48\linewidth]{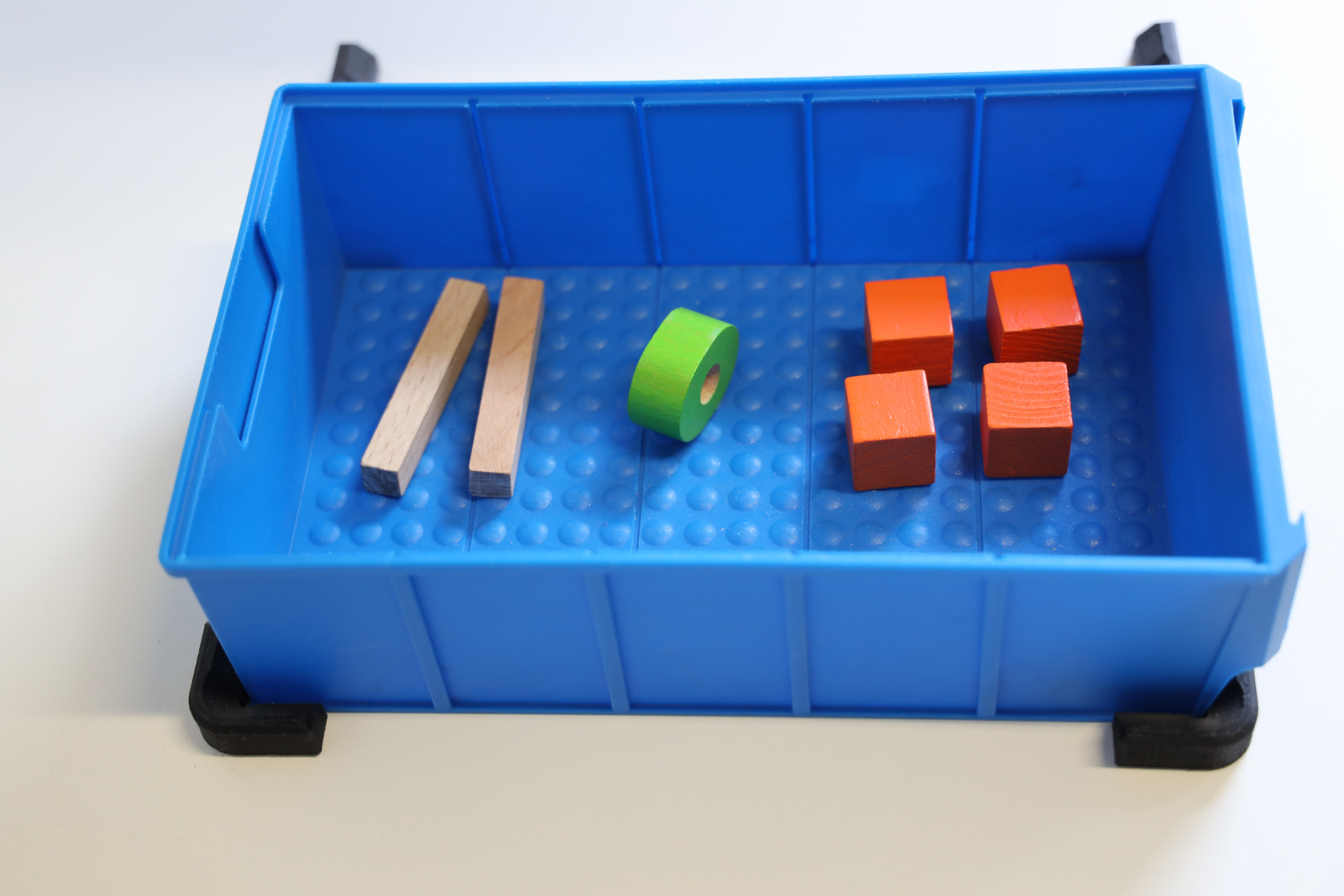}};
        \node[anchor=west,inner sep=0] at (0.1,0) {\includegraphics[trim=180 190 60 100, clip, width=0.48\linewidth]{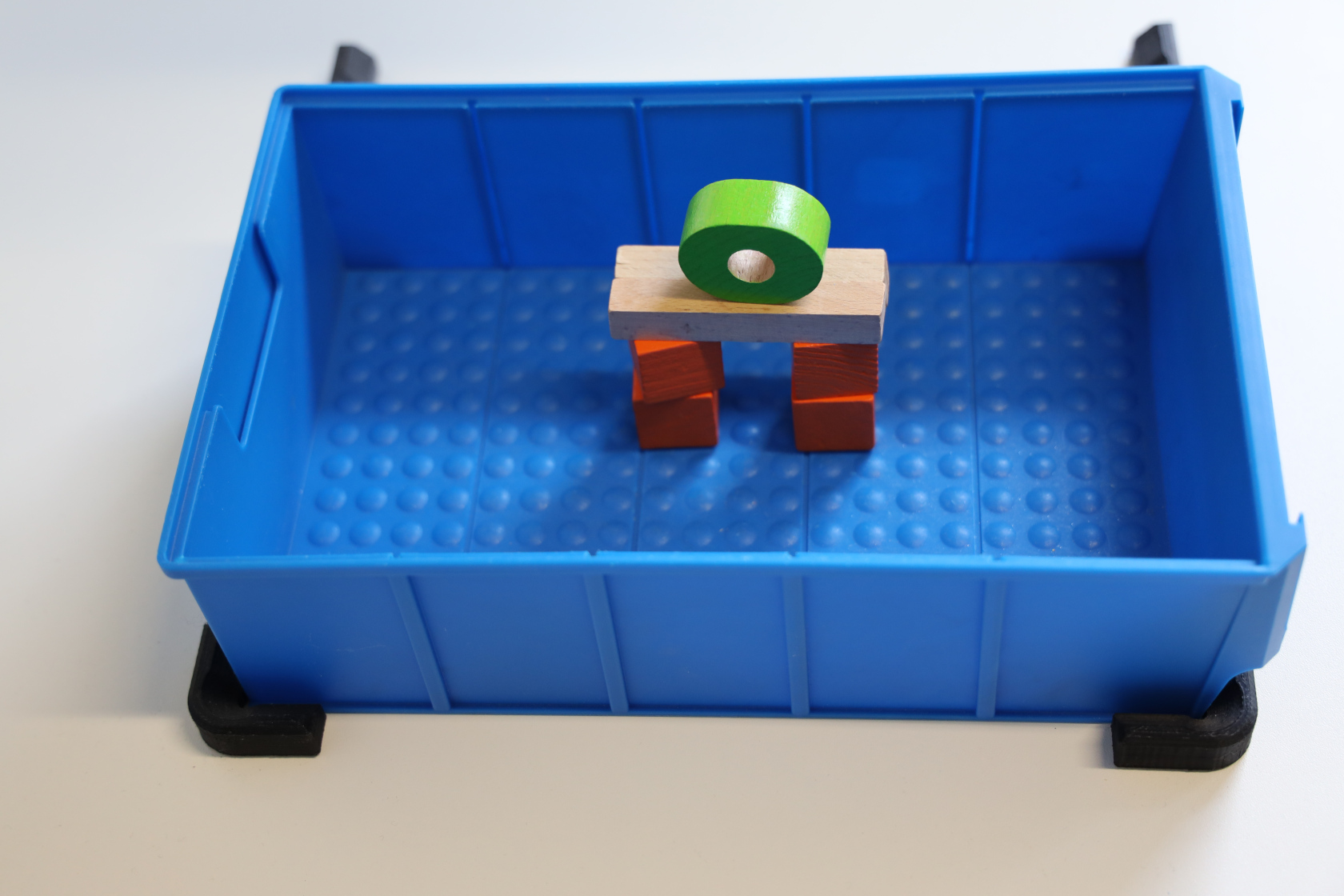}};
        
        \draw[double arrow=6pt colored by black and white] (-0.6,0.2) to [out=30, in=150] (0.7,0.2);
    \end{tikzpicture}
		\label{fig:examples-multiple-steps-building}
	}
	
	\caption{Given a single goal state, our robot is able to infer \textit{multiple} pick-and-place actions from the (cluttered) grasp scene (left) to the place scene (right). Using an instruction list of multiple goal states, the robot is able to reproduce more complex examples (c).}
	\label{fig:examples-multiple-steps}
\end{figure}
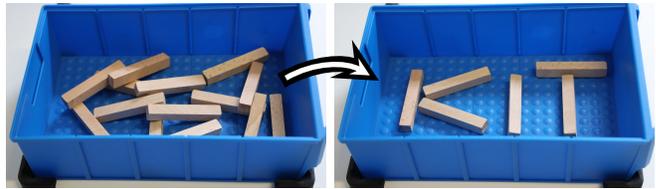
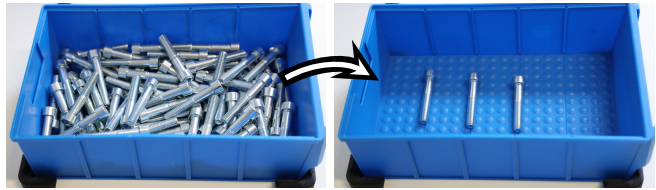
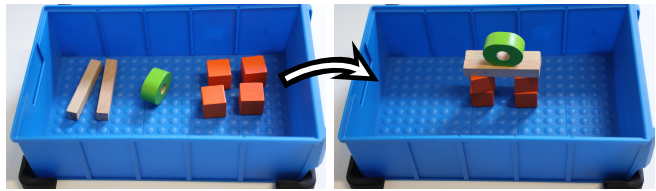

\section{Discussion and Outlook}

We have presented an approach for learning pick-and-place tasks in a self-supervised manner. Since our approach does not depend on an object model and instead takes a demonstrated goal state as an input, it allows for the flexible yet precise placing of unknown objects. The evaluated system was trained in the real-world with up to \num{25000} pick-and-place actions, resulting in average placement errors of \SI{2.7 \pm 0.2}{mm} and \SI{2.6 \pm 0.8}{^\circ} for trained objects. A separated approach, where grasp and place actions are calculated independently, results in four to five times lower precision. For unknown objects, the translational placement error increases to around \SI{6}{mm} and \SI{9}{mm} for grasping in clutter. The robot learns to select the demonstrated objects out of five alternatives with up to \SI{86}{\%} accuracy. A second model was trained specially for screws in an industrial use case. Moreover, we demonstrated the robot's capability to pick-and-place multiple objects from a single goal state. \\

We see our system as a combination of two common approaches in robot learning: First, we build upon learning of manipulation primitives, estimating their reward for planar poses using \acp{FCNN}. So far, we found that this was mostly done for grasping or pre-grasping \cite{zeng2018robotic, mahler_dex-net_2017, zeng2019tossingbot}, as the reward can here be defined and measured more easily. Second, we integrate methods of one-shot imitation learning. In particular, the work of Singh et al.~\cite{singh2019end-to-end} used a contrastive loss approach to classify goal states from policy-generated, self-executed states. Similarly, we use a demonstrated goal state to define a precise object pose for placing.

Regarding other approaches to pick-and-place, we see both advantages as well as shortcomings. In comparison to Finn et al.~\cite{finn2017one}, we limit our policy to a single time step and a discrete action space. While this diminishes the generality and ignores the trajectory in between, our approach increases the final object precision significantly. Moreover, we extend a common, state-of-the-art approach of learning for grasping, i.a.\ resulting in the capability of pick-and-place out of clutter. Due to the use of manipulation primitives, our trained models depend only on the gripper and generalize in principle to other robotic arms.

Gualtieri et al.~\cite{gualtieri2018pick} learn a policy for pick-and-place in full 6-\ac{DoF}. Despite this impressive advantage over our 4-\ac{DoF} planar manipulation, we see shortcomings in the restricted generalization capability as well as a reward-based placing objective. In contrast, our approach allows for wider generalization, easy training of additional object types and flexible place poses. Additionally, our robot is limited by its data consumption of real-world training in comparison to learning in simulation.

As a part of a pick-and-place pipeline, Zhao et al.~\cite{zhao2019towards} predicted the object displacement during the gripper's closing action. We found that this displacement is a major source of imprecision in our experiments. Still, our overall pick-and-place precision is similar to their displacement prediction error. In comparison, our contributions allow to learn the entire pick-and-place pipeline at once. Their approach would still require a pose estimation of the object model and a grasp point detection for targeted placing. Furthermore, our approach was also evaluated in clutter. \\

Regarding this grasp displacement, we assume that our approach is limited by its open-loop nature: Both grasping and placing actions are planned ahead. In future work, we will explore ways to increase the precision of pick-and-place actions, for example by observing the grasped object within the gripper. Then, we can close the loop and refine the place action after picking up the object.

Moreover, we would like to extend our work to pick-and-place in the same scene, e.g.\ for correcting placed objects. Additional manipulation primitives might augment the robot's capability, in particular for placing objects using discrete rotations in full 6-\ac{DoF}. With these ideas in mind, we hope that our research will pave the way to greater flexibility in production and industrial automation.

\section*{Acknowledgement}

We would like to thank Tamim Asfour for his helpful suggestions and discussions.

\bibliographystyle{IEEEtran}
\IEEEtriggeratref{3}
\bibliography{references}

\end{document}